\documentclass[conference]{IEEEtran}
\IEEEoverridecommandlockouts
\usepackage{cite}
\usepackage{amsmath,amssymb,amsfonts}
\usepackage{algorithmic}
\usepackage{graphicx}
\usepackage{multicol,multirow,rotating,booktabs,float}
\usepackage{tabularx}
\usepackage{supertabular}
\usepackage[export]{adjustbox}
\usepackage{subcaption}
\usepackage{textcomp}
\usepackage{fancyhdr}
\usepackage{xcolor}

\def\BibTeX{{\rm B\kern-.05em{\sc i\kern-.025em b}\kern-.08em
    T\kern-.1667em\lower.7ex\hbox{E}\kern-.125emX}}

\begin{document}

\title{DAPAS : Denoising Autoencoder to Prevent Adversarial attack in Semantic Segmentation\\
}

\author{\IEEEauthorblockN{Seungju Cho}
\IEEEauthorblockA{\textit{School of Computing} \\
\textit{KAIST} \\
Daejeon, South Korea \\
joyga@kaist.ac.kr}
\and
\IEEEauthorblockN{Tae Joon Jun}
\IEEEauthorblockA{\textit{Asan Medical Center} \\
Seoul, South Korea \\
taejoon@amc.seoul.kr}
\and
\IEEEauthorblockN{Byungsoo Oh}
\IEEEauthorblockA{\textit{School of Computing} \\
\textit{KAIST} \\
Daejeon, South Korea \\
bsoh@kaist.ac.kr}
\and
\IEEEauthorblockN{Daeyoung Kim}
\IEEEauthorblockA{\textit{School of Computing} \\
\textit{KAIST} \\
Daejeon, South Korea \\
kimd@kaist.ac.kr}
}

\maketitle

\begin{abstract}
Nowadays, deep learning techniques show dramatic performance in computer vision areas, and they even outperform humans on complex tasks such as \textit{ImageNet} classification. But it turns out a deep learning based model is vulnerable to some small perturbation called an adversarial attack. This is a problem in the view of the safety and security of artificial intelligence, which has recently been studied a lot. These attacks have shown that they can easily fool models of image classification, semantic segmentation, and object detection. We focus on the adversarial attack in semantic segmentation tasks since there is little work in this task. We point out this attack can be protected by denoise autoencoder, which is used for denoising the perturbation and restoring the original images. We build a deep denoise autoencoder model for removing the adversarial perturbation and restoring the clean image. 
We experiment with various noise distributions and verify the effect of denoise autoencoder against adversarial attack in semantic segmentation task. 

\end{abstract}

\section{Introduction}

A deep neural network has shown remarkable performance in vision-related tasks such as image classification, object detection, and semantic segmentation. With this performance, deep learning technology has started to be applied to various practice areas such as a self-driving car, health care artificial intelligence.
However, according to a recent study, deep learning models are vulnerable to well-designed perturbation of input. These perturbations are hard to detect via human eyes, so humans can still understand objects correctly. But a deep neural network can produce completely different results than we expect. The adversary can even make perturbation in the way they want. For instance, they can change the image so that a deep neural network misclassifies it as a wrong target set by them. This phenomenon is an important issue in terms of security and safety of artificial intelligence \cite{sitawarin2018darts,eykholt2018robust,finlayson2018adversarial}. 

This perturbed image is called an \textit{adversarial example} or \textit{adversarial attack}. It is generated by using the parameters and loss function of the victim model. And it is called \textit{white box} attack since it requires the information of the model.  
\begin{figure}[!tbh]
    \centering
  \begin{subfigure}{0.4\columnwidth}
  \includegraphics[width=\textwidth]{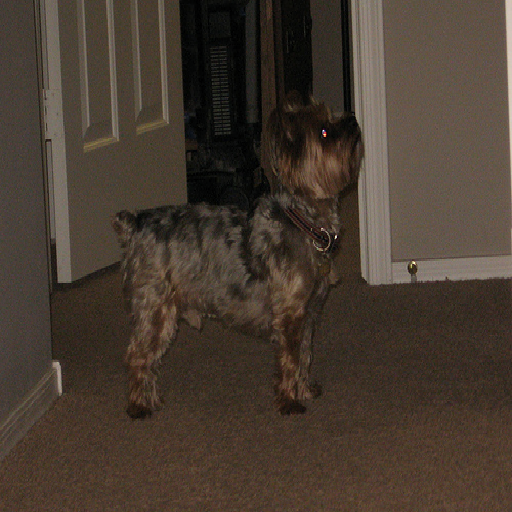}
  \caption{Clean image}
  \end{subfigure}
  \hfill
  \begin{subfigure}{0.4\columnwidth}
  \includegraphics[width=\textwidth]{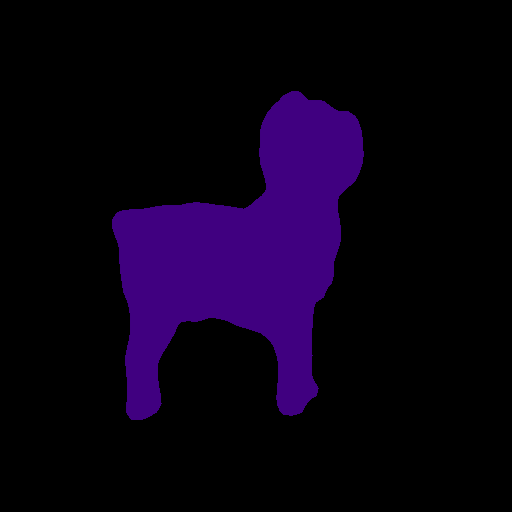}
  \caption{Prediction of (a)} 
  \end{subfigure} 
  \begin{subfigure}{0.4\columnwidth} 
  \includegraphics[width=\textwidth]{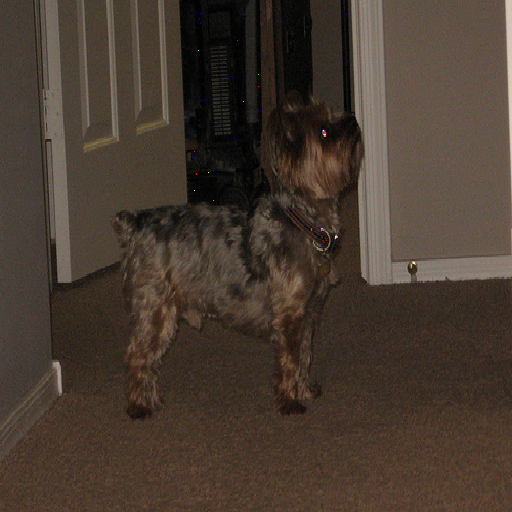}
  \caption{Adversarial example} 
  \end{subfigure}  
  \hfill 
  \begin{subfigure}{0.4\columnwidth} 
  \includegraphics[width=\textwidth]{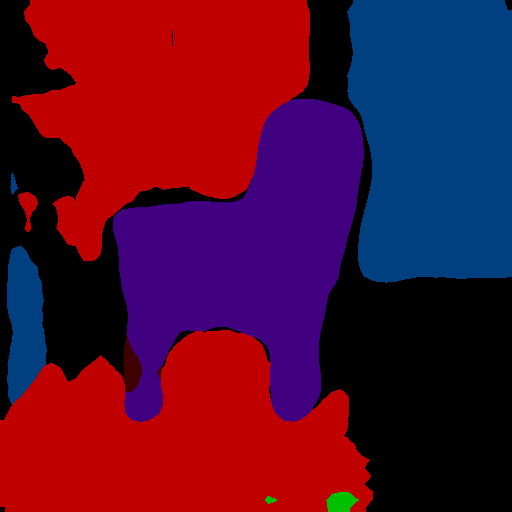} 
  \caption{Prediction of (c)} 
  \end{subfigure}
  \caption{Adversarial attack in semantic segmentation}
  \label{fig:figure1}
  \end{figure}
  But in case of no access to the model, it is still possible to create an adversarial example because of the property called the \textit{transferability}. 
  It allows the attack on this situation, which is called \textit{black box} attack. i.e., an adversarial example generated from a specific model works with other models as well. It is known that an adversarial attack works better with a similar task. For instance, an adversarial attack on a particular neural network can fool other neural networks with different architectures\cite{papernot2016transferability}. With this transferable feature, an adversary can more easily fool diverse models.
  
Many research has tried to generate stronger adversarial examples to attack the state-of-the-art models  \cite{papernot2016limitations,moosavi2016deepfool,carlini2017towards}. And in response to this, there has been some research approaches to make robust models against adversarial attack  \cite{papernot2016distillation,meng2017magnet}. There were also competitions involving adversarial attacks and defense in the field of image classification \cite{kurakin2018adversarial}. Several methods for attack and defense are proposed. However, many of the attacks and defense involved in adversarial examples have been limited to the problem of image classification. Although there are related works which deal with the semantic segmentation and object detection \cite{thys2019fooling,8237562,xie2017adversarial}, only attack scenarios are addressed and studies of defense scenarios for complex tasks such as semantic segmentation are insufficient. Also, for the defense model of the classification task, they often use simple datasets like MNIST, CIFAR 10, that have a small resolution. Therefore, it is needed to experiment with images that have large resolution since more complicated images are used in the real world. In addition, we need to study the defense scenario of semantic segmentation since it is more practical.  For instance, in autonomous driving vehicles or medical intelligence, most of the scene understanding is performed through semantic segmentation rather than classification. 

In this paper, we aim to provide robust mechanisms to secure semantic segmentation model from adversarial attack
To achieve that, we propose DAPAS, a denoise autoencoder to prevent an adversarial attack in semantic segmentation that effectively removes adversarial perturbation.   Since semantic segmentation involves the classification of pixels, it is important to restore the original image at the pixel level so that the restored image gives the correct semantic segmentation result.  We use random noise that follows a particular distribution. We use the Gaussian distribution, Uniform distribution, and Bimodal distribution. The adversarial attack would change the pixel value of input X slightly, the random noise could cover a variety of attack methods. 
For the dataset, we use the PASCAL VOC 2012 \cite{everingham2010pascal} and test it on the DeepLab V3 Plus \cite{chen2018encoder} which has one of the state-of-the-art models in the field of semantic segmentation task. We first generate an adversarial example of DeepLab V3 Plus, and we verify that our approach is effective against adversarial attacks on semantic segmentation. As a result, the performance of our proposed model was around 97 \% compared to the original model DeepLab V3 Plus on clean images. In the case of an adversarial attack, the performance of DeepLab V3 Plus dropped to about 13 \% of the original performance, but when it passed our denoise autoencoder, it covered up to 68 \% compared to the original performance.
Therefore, the method we proposed confirmed that the attack is effectively defended while minimizing performance degradation. In addition, we don't have to retrain the segmentation model. We leave the model we want to defend as it is, and we defend the model by putting a DAPAS in front of it.

The content of the remaining parts is as follows. We show an overall review of related work in Section 2. And we explain our method and show our architecture in Section 3. In Section 4, We evaluate our defense method with an adversarial example. And the conclusion and discussion are provided in Section 5.

\section{Related Work}

Szegedy et al. found the existence of perturbation that breaks the classifier \cite{42503}. This paper presents a simple and effective attack called Fast Gradient Sign Method (FGSM) \cite{43405}. It shows small perturbation is enough to fool the classifier. \cite{moosavi2016deepfool} measures the minimum size required for the attack. They give better intuition of the existence of an adversarial example by calculating the sufficient magnitude of the perturbation. \cite{su2019one} set more extremely limited scenario. They show that the modification of just one pixel of an image would be dropping the performance.  \cite{papernot2017practical} study the scenario where the adversary does not know about the deep learning model, which is called in the black box scenario. This research implies that adversaries can attack even though they have no detail information about the deep learning model. In addition to classification problems, \cite{8237562,xie2017adversarial} shows the adversarial attack on the task of the segmentation model and object detection. And \cite{arnab2018robustness} experiments and analyzes the effect of the adversarial attack on the various semantic segmentation models such as DeepLab V2 \cite{chen2017deeplab} and PSPNet  \cite{zhao2017pyramid}.
In addition to the theoretical aspect, some papers \cite{Eykholt_2018_CVPR,sitawarin2018darts} show that the existence of adversarial examples in the real world. 

To counter adversarial attacks, some works trained the model with a clean example and adversarial example, which is called adversarial training \cite{42503,43405,kurakin2016adversarial,tramer2017ensemble}. During the process of training, they generate adversarial for the training. Although it works, it depends on the particular adversarial data used in the training process. For instance, \cite{kurakin2016adversarial} shows their approach is robust in the simple attack, but not in a more sophisticated attack. In addition, it has an engineering penalty since it requires retraining the model. If it takes longer to create an adversarial example, it will take more time to retrain the model. So it is a burden to use a more complicate algorithm for generating adversarial examples. 
Instead of using the data augmentation, methods to change the model itself were also proposed  \cite{elsayed2018large,cisse2017parseval}. They change the objective function of the problem for obtaining the robustness. However, this approach also requires retraining the model so it costs time. \cite{samangouei2018defense,meng2017magnet} preprocess the image before putting it into the model. These approaches are similar to our work, but they experiment with the image which has a small resolution like MNIST and CIFAR series. Also, theses work only focus on the classification problem.

Currently, there is no general defense method of adversarial attacks. In addition, to the best of our knowledge, there is no defense scenario of the semantic segmentation in the context of the adversarial attack. We verify our approach is effective against an adversarial attack in semantic segmentation.

\begin{figure*}
\begin{center}
    \includegraphics[width=1\textwidth]{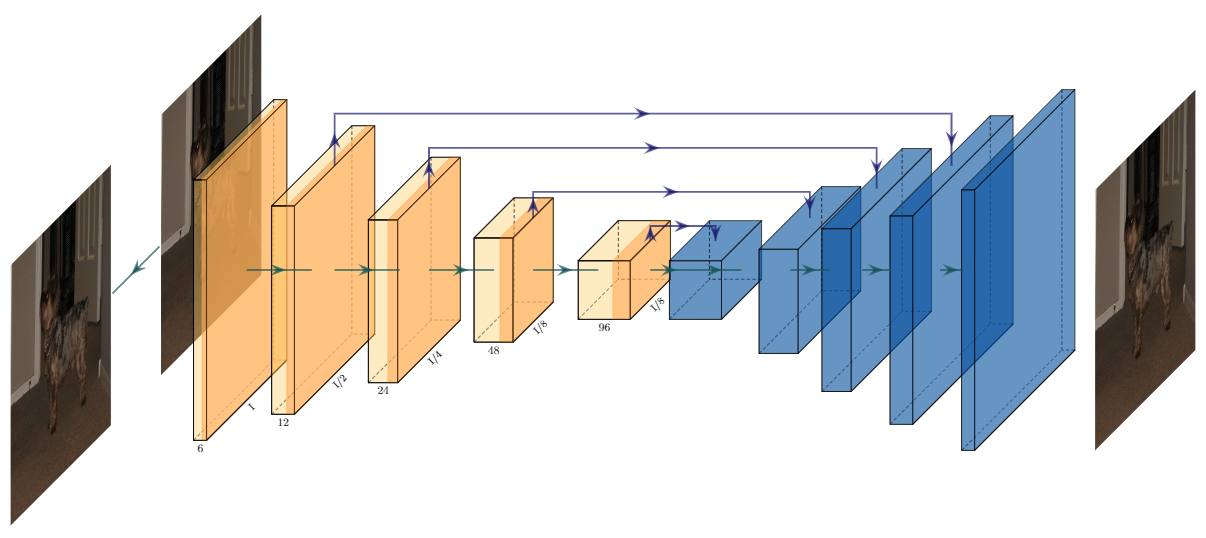}
\end{center}
   \caption{The architecture of the denoise autoencoder }
\label{fig:figure2}
\end{figure*}

\section{Adversarial Attack}

Basically, all of the attacks use the gradient of data with respect to the loss function of the victim model.
 In this section, we briefly review the basic method of adversarial attack.  
\subsection{Fast Gradient Sign Method (FGSM)}
 The FGSM is proposed by \cite{43405}. It is a simple and effective attack method. The image $X$ is perturbed as follows.
$$ X = X + \epsilon \cdot  | sign(\triangledown_{X} l(X,y_{true})) | $$
Where $\epsilon$ is a magnitude of noise and $l(X,y_{true})$ is a loss with respect to the true label of the image.
It adjusts input X by adding a sign of the gradient of X. It increases the loss function of the victim model so that the model misjudges the adjusted input. In other words, it uses a gradient ascent algorithm for increasing the loss value.
Since it updates input X once, it is also called \textit{single-step} method. In addition, Adversaries can update the input in the direction they want. It decreases the loss of the victim model with respect to the target label set by the adversary. 
$$ X = X - \epsilon \cdot   | sign(\triangledown_{X} l(X,y_{target})) | $$
If the input is modified enough, the model predicts the target which the adversary wants. In both cases, the $\epsilon$
have the role of the scale of the perturbation. We call the first method as untargeted FGSM and second method as targeted FGSM.
\subsection{Iterative FGSM (I-FGSM)} The iterative FGSM is a repetitive version of FGSM. it is a more powerful attack method compared to the FGSM. It uses the following equations:
\begin{flalign*}
   X_{t+1} &= clip_{X,\epsilon}(X_{t} + \alpha  \cdot   | sign(\triangledown_{X} l(X_{t},y_{true})) |)
\end{flalign*}
Where   $X_{0} = X$, $\alpha$ is a step size for adjusting $X_t$ , and clip function ensures that $X_t \in (X-\epsilon,X+\epsilon )$ for all $t$. And we choose step number as $ min (\epsilon + 2, 4\epsilon$) if $\epsilon \leq 0.008$, otherwise $min (\epsilon + 4, 1.24\epsilon)$.
It is also called \textit{multi-step} method. Here $\alpha$ is the $1$ on the scale of 0 to 255 in the original paper. We use 0.25 for the $\alpha$. 
As the case of targeted FGSM, the adversary can modify the data in the way they want.
 $$ X_{t+1} = clip_{X,\epsilon}(X_{t} -  \alpha \cdot   | sign(\triangledown_{X} l(X,y_{target}))|)$$
 We denote this algorithm I-FGSM in this paper. Although theses attack methods were introduced in the context of image classification, the same method can be applied in the context of semantic segmentation tasks. We call the first method as an untargeted I-FGSM and second method as targeted I-FGSM. We use untargeted FGSM and untargeted I-FGSM in this paper. i.e., the pixel in the image will be misclassified randomly by an adversarial attack.


\section{Method}
 Our mechanism does not modify the semantic segmentation model. We use denoise autoencoder as a preprocessor. And we place it in front of the semantic segmentation model. Whether the input is an adversarial example or not, it will pass the denoise autoencoder before passing the original model. Thus denoise autoencoder should restore the clean image well, in addition to remove the perturbation. In this section, we show the architecture of denoise autoencoder, detail of training setup and demonstrate how the denoise autoencoder is deployed in constructing a robust semantic segmentation model.

\subsection{Architecture of Denoise Autoencoder}

The overall structure of the model is shown in Fig. \ref{fig:figure2}. The denoise autoencoder is divided into two parts,  orange-colored encoder part, and blue-colored decoder part. While the encoder extracts the feature of the input image, the decoder restores the input image from the extracted feature. The encoder consists of five convolutional layers and the resolution of each feature map gradually reduces by half. We did not use max pooling or average pooling for decreasing the resolution of the feature map. We adjust the stride of the kernel for decreasing the resolution. In the decoder, it also consists of five deconvolutional layers and the resolution of each feature map expands twice. And we use skip connection to restore the details of the spatial information of the feature maps. In other words, the features used in the encoder were symmetrically linked to the features of the decoder. We add features that have the same resolution. Here we did not connect the first feature through the skip connection, i.e. input image.  Since the input image has noisy information. For the activation function, we use ELU instead of RELU for each layer. And we use sigmoid as the last activation function.

\begin{figure*}[!tbh]
	\centering
	\includegraphics[width=1\linewidth]{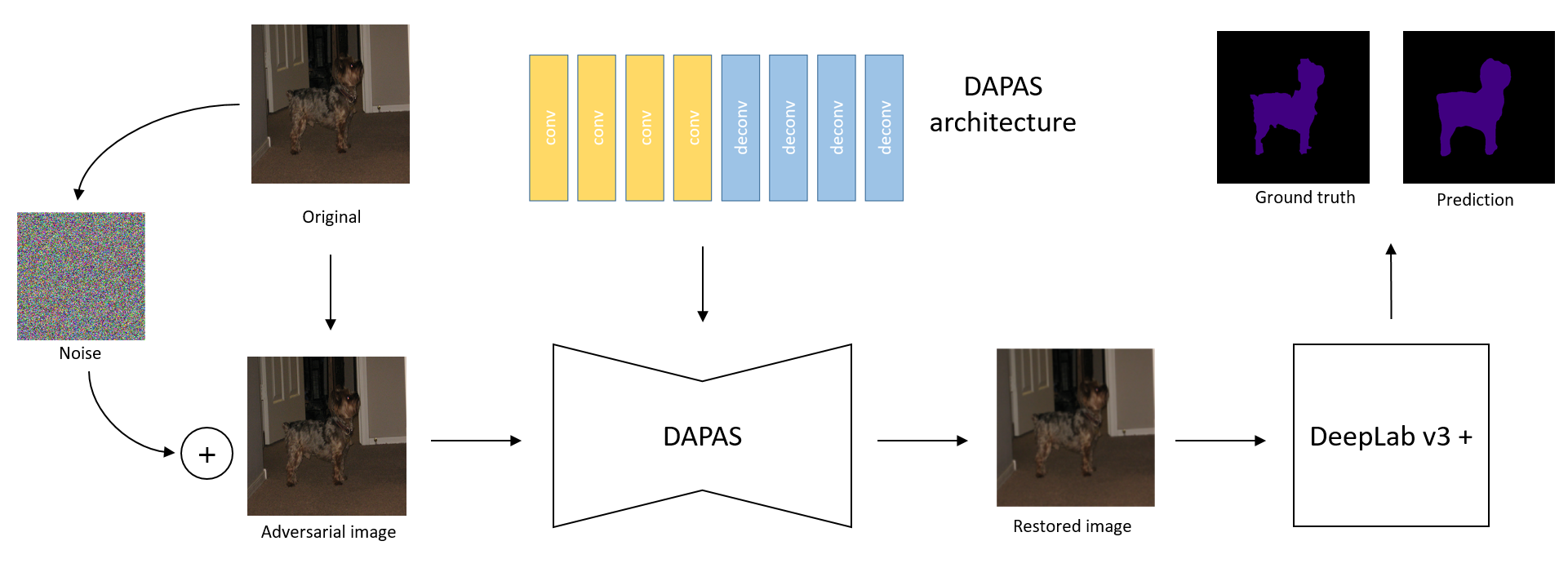}
    \caption{Overall flow of DAPAS}
    \label{fig:figure3}
\end{figure*}

\subsection{Training}
 We trained with PASCAL VOC 2012 data which is
widely used for the task of semantic segmentation. There
are a total of 1464 training images and 1449 validation
data. The pixel value with a value from 0 to 255 was
re-scaled to change from 0 to 1. And the resolution of the
image was fixed to 553 $\times$ 553.
 We use Adam optimizer for gradient descent algorithm, and use $5 \times \ 10^{-4}$ for the learning rate. We add random noise to the train set. And we use either clean or noisy input. It is a little bit different from the original framework of denoise autoencoder \cite{vincent2010stacked}. Since our purpose is to also maintain the original performance in case of no adversarial attack, we also put clean input. For random noise, we use Gaussian distribution, Uniform distribution, and Bimodal distribution. 
For the Gaussian distribution, we set a mean of zero and the standard deviation of 0.004. For the Uniform distribution, we set the range from -0.035 to 0.035. For the Bimodal distribution, we use a mixture of two Gaussian distributions. For each Gaussian distribution, we set a mean of -0.024 and 0.024. And we use the standard deviation of 0.004.

\subsection{Combining with semantic segmentation model}
The created denoise autoencoder is connected to the general model that performs semantic segmentation like Fig. \ref{fig:figure3}. The denoise autoencoder has the role of preprocessing before the image entering the segmentation model. Since the denoise autoencoder is independent of the segmentation model, it can be located in front of any model. Therefore it serves as a general defense mechanism. 
Hence, we do not have to re-train the model we want to defend. Besides, the random noise used in training the denoise autoencoder is independent of any adversarial attack, it can defend against a variety of attacks.

\section{Experiment}

In this section, we look at the ability of the denoise autoencoder to restore and then measure how the restored image performs in the segmentation model DeepLab V3 Plus.
We test the results of segmentation in DeepLab V3 Plus using test data from the Pascal VOC with additional annotation from SBD \cite{hariharan2011semantic}. In an adversarial attack, we assumed that noise is not large. Since the purpose of the adversarial attack is not to deceive people but to deceive models. So we limit the magnitude of the noise to 0.032 of pixel level. i.e., it changes 3.2 \% of the original pixel value. We properly adjust the standard deviation of Gaussian distribution and Bimodal distribution and the range of Uniform distribution. The details of the distribution, the evaluation metric, and the result of the experiment are summarized below.


\subsection{Dataset}
We use the PASCAL VOC validation set which is not used in the process of training the denoise autoencoder. And we resize the image as $513 \times 513.$ Since we use a pre-trained model, which uses the normalized data so we did the same for testing. In other words, the average was subtracted and divided by the standard deviation.

\subsection{Denoise Autoencoder as a restoration}
\begin{figure*}[!tbh]
  \begin{subfigure}[t]{0.5\columnwidth}
  \includegraphics[width=0.9\textwidth,valign=t]{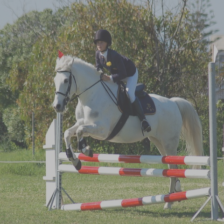}
  \caption{Clean image}
  \end{subfigure}
  \hfill
  \begin{subfigure}[t]{0.5\columnwidth}
  \includegraphics[width=0.9\textwidth,valign=t]{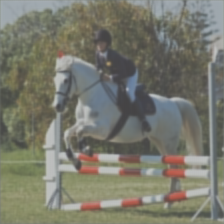}
  \caption{After (a) passing the denoise autoencoder} 
  \end{subfigure} 
  \begin{subfigure}[t]{0.5\columnwidth} 
  \includegraphics[width=0.9\textwidth,valign=t]{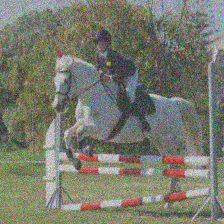}
  \caption{Noisy image} 
  \end{subfigure}  
  \hfill 
  \begin{subfigure}[t]{0.5\columnwidth} 
  \includegraphics[width=0.9\textwidth,valign=t]{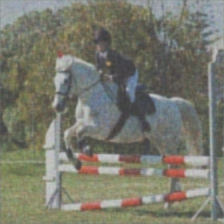} 
  \caption{After noisy image passing the denoise autoencoder} 
  \end{subfigure}
  \caption{Effects of denoise autoencoder}
  \label{fig:figure4}
  \end{figure*}

We visualize the output of denoise autoencoder before measuring the robustness against adversarial attacks. Although we use three different distribution for noise, we show the case of Gaussian noise for simplicity.  Since the noise level is not much different, the results of other distributions are similar. We make sure that the clean image is well restored after denoise autoencoder as well as in case of the noisy image since the performance on a clean image should not be compromised. Fig. \ref{fig:figure4}-(a) shows the original image, Fig. \ref{fig:figure4}-(b) shows the noisy image, Fig. \ref{fig:figure4}-(c) shows the original image after denoise autoencoder and Fig. 11 shows the noisy image after denoise autoencoder. It is easy to see that Fig. \ref{fig:figure4}-(c) is more clear than Fig. \ref{fig:figure4}-(d). Therefore we can intuitively expect that the reduction ratio will be not that much. 

\subsection{Evaluation metric}
The mean Intersection over Union (mIoU) is widely used for evaluating the performance of semantic segmentation \cite{everingham2010pascal}. And we adapt relative metric \textit{IoU Ratio} for measuring the robustness \cite{arnab2018robustness}. The IoU ratio on the attack is defined as follows.
\begin{itemize}
      \item ${mIoU_{CO}}$ :  mIoU of a clean image on the original\\ \hphantom \quad \quad \quad  \quad \enspace model 
      \item ${mIoU_{AO}}$ :  mIoU of an adversarial image on the \\ \hphantom \quad \quad \quad  \quad \enspace original model 
      \item ${mIoU_{CP}}$ :  mIoU of a clean image on the proposed \\ \hphantom \quad \quad \quad  \quad \enspace model 
      \item ${mIoU_{AP}}$ :  mIoU of an adversarial image on the \\
      \hphantom \quad \quad \quad  \quad \enspace  proposed model 
\end{itemize}
$$ IoU \ ratio \ of \ attack \ (RatioATT) = \frac{mIoU_{AO}}{mIoU_{CO}} $$
$$ IoU \ ratio \ of \ robust \ (RatioROB) = \frac{mIoU_{AP}}{mIoU_{CO}} $$

This is a metric that shows the performance is compared to the original model performance. And we measure the mIoU ratio of the original model to the proposed model to calculate the reduction on a clean image as the following.
$$ IoU \ ratio \ of \ reduction \ (RatioRED) = \frac{mIoU_{CP}}{mIoU_{CO}} $$

\subsection{Analysis of results}
Table \ref{table:table1} shows that the performance reduction due to the denoise autoencoder was as low as around 3\% for all three distributions. And we verify that denoise autoencoder is effective to remove an adversarial attack. Fig. 6 show the results. We can see that the segmentation output is weird when the FGSM and the I-FGSM are applied. In addition, We can check that the I-FGSM is less noisy than the FGSM, but the segmentation results show that the attack was more effective in I-FGSM. However, after passing the denoise autoencoder, the images are successfully purified in both cases. Table \ref{table:table2} shows \textit{IoU ratio of attack}. When the $\epsilon$ is 0.008, 0.0016 and 0.0032, \textit{IoU ratio of attack} on FGSM are similar. But in the case of I-FGSM, \textit{IoU ratio of attack} significantly drops to 24.2\%, 21.9\%, and 12.0\%. Table \ref{table:table3} and Table \ref{table:table4} show \textit{IoU ratio of robust} on FGSM and I-FGSM. Comparing the two tables, \textit{IoU ratio of robust} on I-FGSM is larger than \textit{IoU ratio of robust} on FGSM although the attack is more effective on I-FGSM. Fig. \ref{fig:graph_FGSM} and Fig. \ref{fig:graph_IFGSM} summarized the contents of the tables.


\textbf{Gaussian distribution}
We use the small standard deviation of Gaussian distribution since we want to check the performance depends on the noise distribution, so we use the mean of 0 and the standard deviation of 0.004. Although it is also effective, the performance is middle among the three denoise autoencoders.
    
\textbf{Uniform distribution}
We give the range from -0.035 to 0.035 of Uniform distribution since the maximum magnitude of adversarial perturbation is 0.032. Among the three distribution, denoise autoencoder using the Uniform distribution has the best performance in terms of reduction. It shows a 97.4\% IoU ratio of reduction. Therefore, it only decreased by 3.1 \% compared to the original performance. In the case of the attack scenario, it shows the worst performance compared to other distributions.
    
\textbf{Bimodal distribution}
In most cases, denoise autoencoder trained with Bimodal distribution noise shows the best performance among the three distribution. Since the adversarial attack would add or subtract noise of fixed size, we use the noise following the Bimodal distribution using the two Gaussian distributions like Fig. 5. Each Gaussian distribution has a mean of  -0.24 and 0.24, the standard deviation of 0.004. 
The IoU reduction is 97.2 \%, which is the worst among the three distributions, but the difference is small. And in case of an attack, the performance was the best.

\begin{figure}[H]
\begin{center}
    \includegraphics[width=0.4\textwidth,height = 3cm]{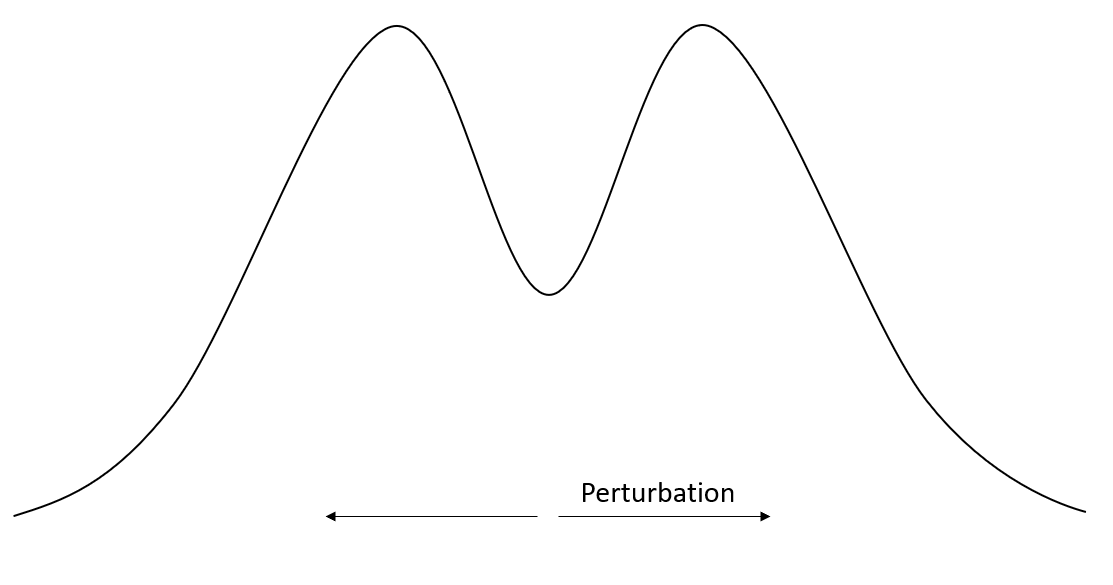}\par
    \caption{Bimodal distribution using two Gaussian distributions}
\end{center}
\label{fig:bimodal}
\end{figure}

\begin{figure*}[h]
    \centering
    \begin{subfigure}[c]{0.2\linewidth}
    \includegraphics[height = 3.8cm,width=3.8cm]{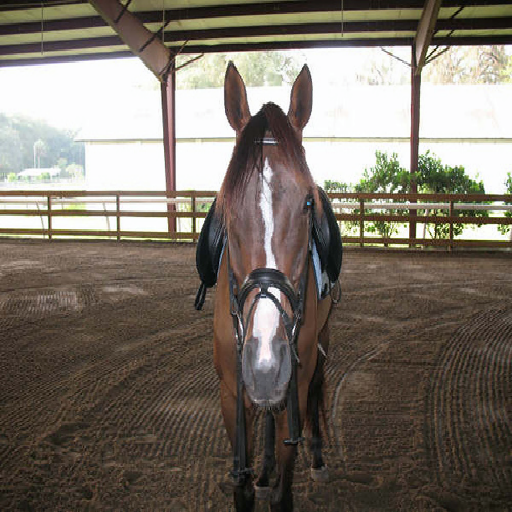}\par
    \caption{Clean image}
    \end{subfigure}
    \hfill
    \begin{subfigure}[c]{0.2\linewidth}
    \includegraphics[height = 3.8cm,width=3.8cm]{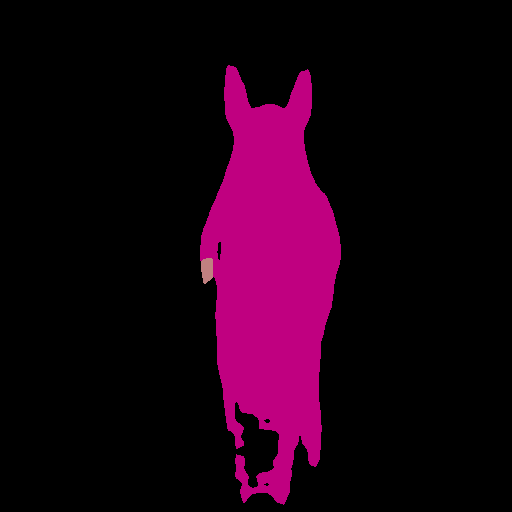}\par
    \caption{Prediction of (a)}
    \end{subfigure}
    \hfill
    \begin{subfigure}[c]{0.2\linewidth}
    \includegraphics[height = 3.8cm,width=3.8cm]{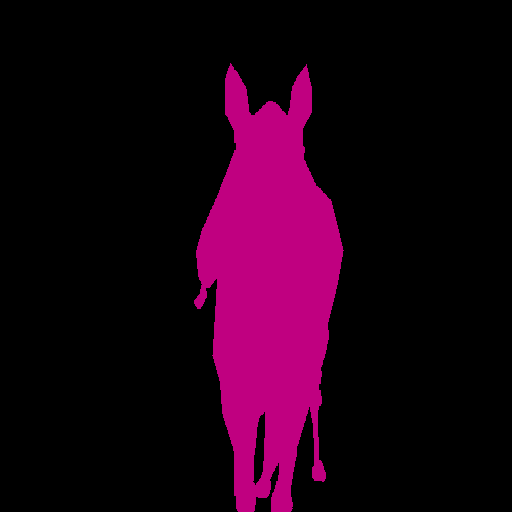}\par
    \caption{Ground truth}
    \end{subfigure}

    \begin{subfigure}[t]{0.2\linewidth}
    \includegraphics[height = 3.8cm,width=3.8cm]{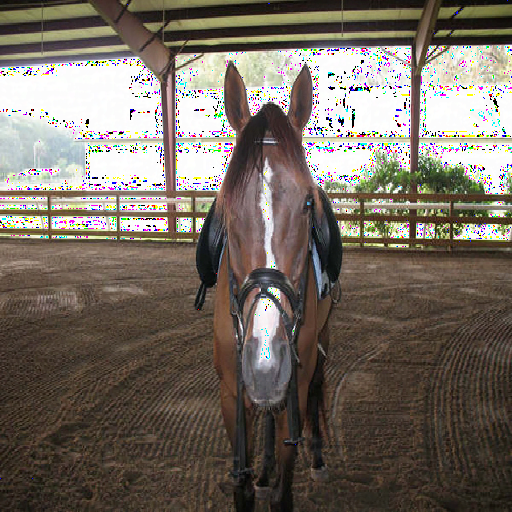}\par
    \caption{adversarial example using FGSM}
    \end{subfigure}
    \hfill
    \begin{subfigure}[t]{0.2\linewidth}
    \includegraphics[height = 3.8cm,width=3.8cm]{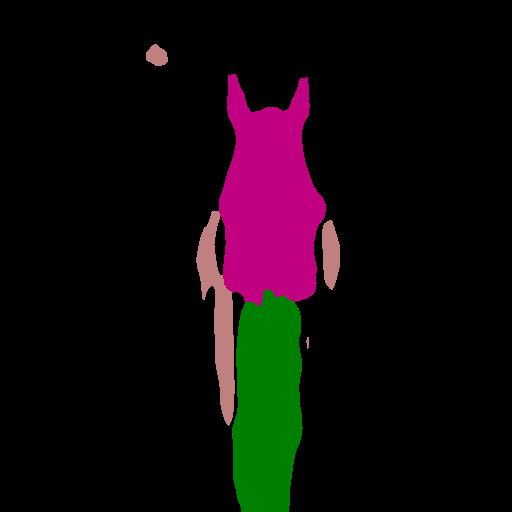}\par
    \caption{Prediction of the FGSM}
    \end{subfigure}
    \hfill
    \begin{subfigure}[t]{0.2\linewidth}
    \includegraphics[height = 3.8cm,width=3.8cm]{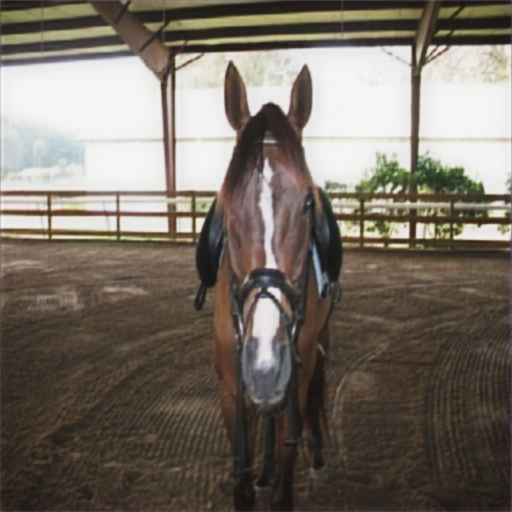}\par
    \caption{After pass the DAPAS}
    \end{subfigure}
    \hfill
    \begin{subfigure}[t]{0.2\linewidth}
    \includegraphics[height = 3.8cm,width=3.8cm]{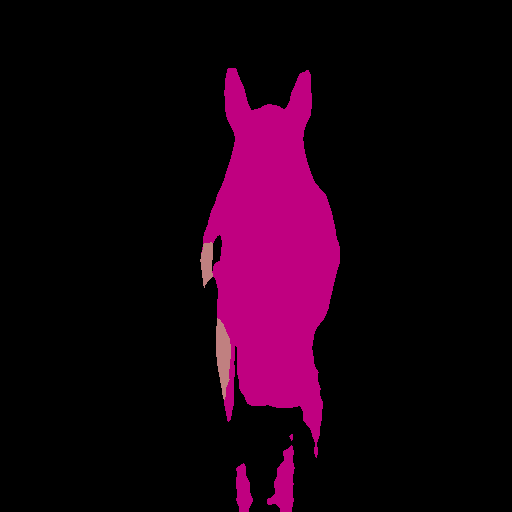}\par
    \caption{Prediction of the (f)}
    \end{subfigure}

    \begin{subfigure}[t]{0.2\linewidth}
    \includegraphics[height = 3.8cm,width=3.8cm]{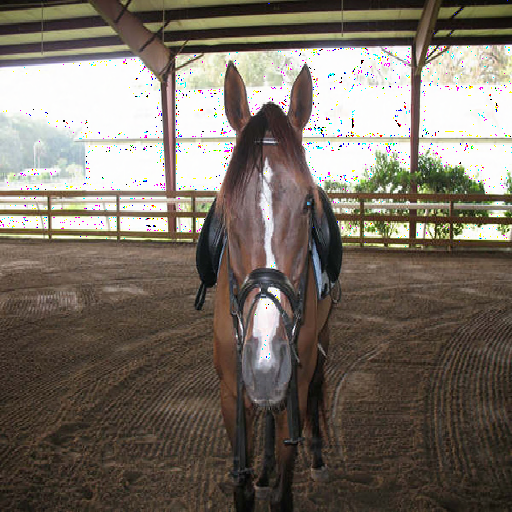}\par
    \caption{adversarial example using I-FGSM}
    \end{subfigure}
    \hfill
    \begin{subfigure}[t]{0.2\linewidth}
    \includegraphics[height = 3.8cm,width=3.8cm]{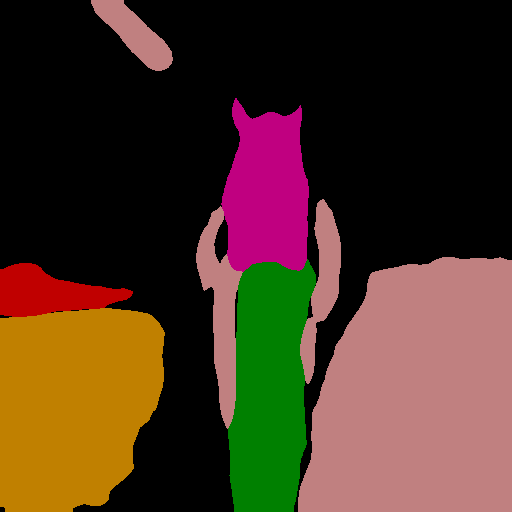}\par
    \caption{Prediction of the I-FGSM}
    \end{subfigure}
    \hfill
    \begin{subfigure}[t]{0.2\linewidth}
    \includegraphics[height = 3.8cm,width=3.8cm]{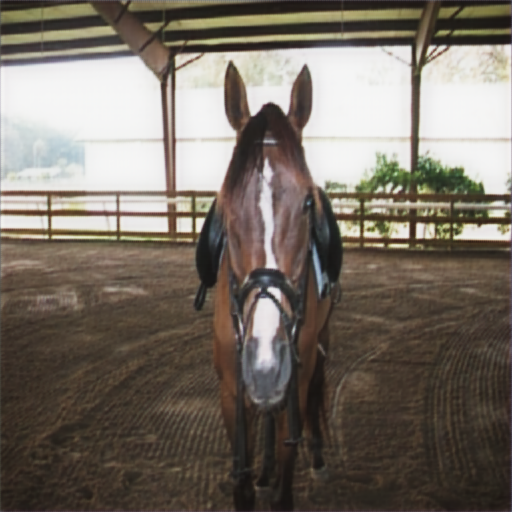}\par
    \caption{After pass the DAPAS}
    \end{subfigure}
    \hfill
    \begin{subfigure}[t]{0.2\linewidth}
    \includegraphics[height = 3.8cm,width=3.8cm]{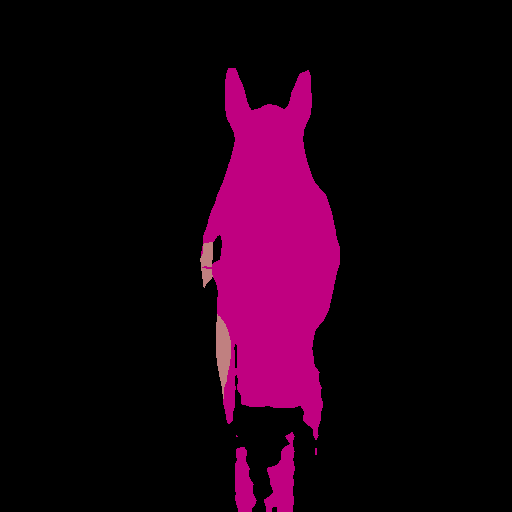}\par
    \caption{Prediction of the (j)}
    \end{subfigure}
    \caption{Images from adversarial attack by using FGSM and I-FGSM, images after DAPAS , and outputs for each.}
\end{figure*}

 \begin{figure*}[!h]
\begin{multicols}{2}
\begin{center}
    \includegraphics[height = 7cm,width=8cm]{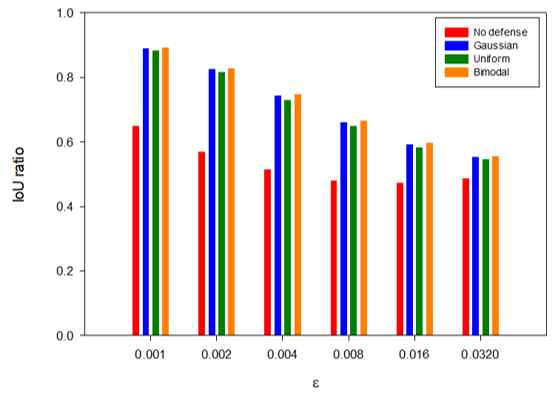}\par
    \caption{IoU ratio of attack on FGSM}
    \label{fig:graph_FGSM}
\end{center}
\begin{center}
    \includegraphics[height = 7cm, width=8cm]{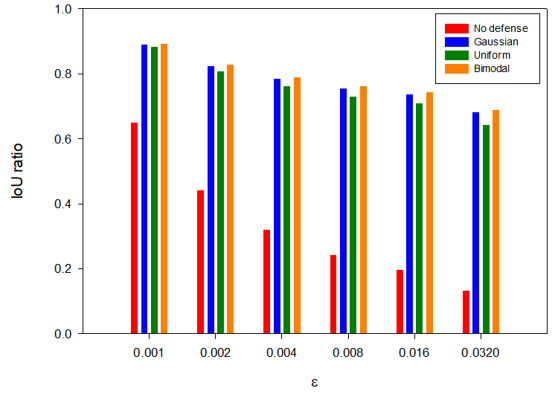}\par
    \caption{IoU ratio of attack on I-FGSM}
        \label{fig:graph_IFGSM}
\end{center}
\end{multicols}
\end{figure*}

\begin{table*}[!h]
    \begin{minipage}{.45\linewidth}
       \centering
           \caption{\textit{IoU ratio of reduction}}
     \scalebox{1}{
    \begin{tabular}{c c c c}
        \toprule
        \midrule
        \multirow{2}[4]{*}{Noise} & \multicolumn{1}{c}{Clean image} &
        \multicolumn{2}{c}{After DAPAS}\\ 
        \cmidrule(rl){2-4}
        & mIoU & mIoU & \textit{RatioRED}(\%) \\ 
        \cmidrule(r){1-1}\cmidrule(l){2-4}
        \multicolumn{1}{c}{Gaussian}&\multirow{3}{*}{78.4} & 76.4 & 97.4  \\
        \multicolumn{1}{c}{Uniform}& & 76.4 & 97.4  \\
        \multicolumn{1}{c}{Bimodal}& & 76.3 & 97.2  \\
        \midrule
        \bottomrule
    \label{table:table1}
    \end{tabular}
    }
    \end{minipage}%
    \begin{minipage}{.6\linewidth}
      \centering
          \caption{\textit{IoU ratio of attack}}
      \scalebox{1}{
        \begin{tabular}{c c c c c }
            \toprule
            \midrule
            \multirow{2}[4]{*}{$\epsilon$} & \multicolumn{2}{c}{FGSM} & \multicolumn{2}{c}{I-FGSM}\\ 
            \cmidrule(rl){2-5}
            & mIoU  & \textit{RatioATT} (\%) & mIoU & \textit{RatioATT} (\%) \\ 
            \cmidrule(r){1-1}\cmidrule(l){2-5}
            \multicolumn{1}{c}{0.001}& 50.9 & 64.8 & 50.9 & 64.8  \\
            \multicolumn{1}{c}{0.002}& 44.6 & 56.9 & 34.5 & 44.0  \\
            \multicolumn{1}{c}{0.004}& 40.2 & 51.2 & 24.9 & 31.8  \\
            \multicolumn{1}{c}{0.008}& 37.6 & 48.0 & 19.0 & 24.2   \\
            \multicolumn{1}{c}{0.016}& 37.1 & 47.3 & 15.3 & 19.6   \\
            \multicolumn{1}{c}{0.032}& 38.0 & 48.5 & 10.3 & 13.1   \\
            \midrule
            \bottomrule
             \label{table:table2}
        \end{tabular}
    }
    \end{minipage} 
\end{table*}

\begin{table*}[!h]
    \centering
        \caption{\textit{IoU ratio of robust} on FGSM}
    \scalebox{1.05}{
    \begin{tabular}{c c c c c c c c c }
        \toprule
        \midrule
        \multirow{2}[4]{*}{$\epsilon$} & \multicolumn{2}{c}{Gaussian} & \multicolumn{2}{c}{Uniform} & \multicolumn{2}{c}{Bimodal} \\ 
        \cmidrule(rl){2-7}
        & mIoU  & \textit{RatioROB} (\%) &mIoU & \textit{RatioROB}  (\%) & mIoU  & \textit{RatioROB} (\%)\\ 
        \cmidrule(r){1-1}\cmidrule(l){2-7}
        \multicolumn{1}{c}{0.001}& 69.7 & 88.9 & 69.2 & 88.2 & 69.8 & 89.1   \\
        \multicolumn{1}{c}{0.002}& 64.8 & 82.6 & 63.9 & 81.5 & 64.9 & 82.8   \\
        \multicolumn{1}{c}{0.004}& 58.3 & 74.3 & 57.2 & 72.9 & 58.6 & 74.7 \\
        \multicolumn{1}{c}{0.008}& 51.8 & 66.0 & 50.8 & 64.8 & 52.0 & 66.4   \\
        \multicolumn{1}{c}{0.016}& 46.5 & 59.2 & 45.7 & 58.2 & 46.7 & 59.6   \\
        \multicolumn{1}{c}{0.032}& 43.3 & 55.2 & 42.7 & 54.5 & 43.4 & 55.3  \\
        \midrule
        \bottomrule
     \label{table:table3}
    \end{tabular}}
    \label{tab:sam_count2}
    \end{table*}

 \begin{table*}[!h]
    \centering
        \caption{\textit{IoU ratio of robust} on I-FGSM}
    \scalebox{1.05}{
    \begin{tabular}{c c c c c c c c c }
        \toprule
        \midrule
        \multirow{2}[4]{*}{$\epsilon$} & \multicolumn{2}{c}{Gaussian} & \multicolumn{2}{c}{Uniform} & \multicolumn{2}{c}{Bimodal} \\ 
        \cmidrule(rl){2-7}
        & mIoU  & \textit{IoU ratio of robust} (\%) & mIoU & \textit{IoU ratio of robust} (\%) & mIoU  & \textit{IoU ratio of robust} (\%)\\ 
        \cmidrule(r){1-1}\cmidrule(l){2-7}
        \multicolumn{1}{c}{0.001}& 69.7 & 88.9 & 69.2  & 88.2 & 69.8 &89.1    \\
        \multicolumn{1}{c}{0.002}& 64.5 & 82.2 & 63.3 & 80.8 & 64.8 & 82.6    \\
        \multicolumn{1}{c}{0.004}& 61.4 & 78.3 &59.7  &76.1  & 61.7 & 78.7 \\
        \multicolumn{1}{c}{0.008}& 59.1 &75.4  &57.1  & 72.9 &59.7  & 76.1  \\
        \multicolumn{1}{c}{0.016}& 57.7& 73.6 &55.5  & 70.8 & 58.3 & 74.3  \\
        \multicolumn{1}{c}{0.032}& 53.3&68.0 & 50.2 &  64.& 53.9 & 68.7    \\
        \midrule
        \bottomrule
     \label{table:table4}
    \end{tabular}}
    \label{tab:sam_count2}
    \end{table*} 
   
\section{Conclusion}

We verify the denoise autoencoder is effective in defending against the adversarial attack in the context of the semantic segmentation task. We also confirm that the performance varies slightly depending on what kind of noise distribution the denoise autoencoder produces in the input. We also believe that since our denoise autoencoder is independent of a particular attack when designing the denoise autoencoder, this approach is available not only in the semantic segmentation task but also in the areas of classification and object detection. The design of a more detailed and careful denoise autoencoder against the adversarial attack remains a future study.

{\small
\bibliographystyle{ieee}
\bibliography{egbib}

\begin{thebibliography}{10}\itemsep=-1pt

\bibitem{arnab2018robustness}
A.~Arnab, O.~Miksik, and P.~H. Torr.
\newblock On the robustness of semantic segmentation models to adversarial
  attacks.
\newblock In {\em Proceedings of the IEEE Conference on Computer Vision and
  Pattern Recognition}, pages 888--897, 2018.

\bibitem{carlini2017towards}
N.~Carlini and D.~Wagner.
\newblock Towards evaluating the robustness of neural networks.
\newblock In {\em 2017 IEEE Symposium on Security and Privacy (SP)}, pages
  39--57. IEEE, 2017.

\bibitem{chen2017deeplab}
L.-C. Chen, G.~Papandreou, I.~Kokkinos, K.~Murphy, and A.~L. Yuille.
\newblock Deeplab: Semantic image segmentation with deep convolutional nets,
  atrous convolution, and fully connected crfs.
\newblock {\em IEEE transactions on pattern analysis and machine intelligence},
  40(4):834--848, 2017.

\bibitem{chen2018encoder}
L.-C. Chen, Y.~Zhu, G.~Papandreou, F.~Schroff, and H.~Adam.
\newblock Encoder-decoder with atrous separable convolution for semantic image
  segmentation.
\newblock In {\em Proceedings of the European conference on computer vision
  (ECCV)}, pages 801--818, 2018.

\bibitem{cisse2017parseval}
M.~Cisse, P.~Bojanowski, E.~Grave, Y.~Dauphin, and N.~Usunier.
\newblock Parseval networks: Improving robustness to adversarial examples.
\newblock In {\em Proceedings of the 34th International Conference on Machine
  Learning-Volume 70}, pages 854--863. JMLR. org, 2017.

\bibitem{elsayed2018large}
G.~Elsayed, D.~Krishnan, H.~Mobahi, K.~Regan, and S.~Bengio.
\newblock Large margin deep networks for classification.
\newblock In {\em Advances in neural information processing systems}, pages
  842--852, 2018.

\bibitem{everingham2010pascal}
M.~Everingham, L.~Van~Gool, C.~K. Williams, J.~Winn, and A.~Zisserman.
\newblock The pascal visual object classes (voc) challenge.
\newblock {\em International journal of computer vision}, 88(2):303--338, 2010.

\bibitem{eykholt2018robust}
K.~Eykholt, I.~Evtimov, E.~Fernandes, B.~Li, A.~Rahmati, C.~Xiao, A.~Prakash,
  T.~Kohno, and D.~Song.
\newblock Robust physical-world attacks on deep learning visual classification.
\newblock In {\em Proceedings of the IEEE Conference on Computer Vision and
  Pattern Recognition}, pages 1625--1634, 2018.

\bibitem{Eykholt_2018_CVPR}
K.~Eykholt, I.~Evtimov, E.~Fernandes, B.~Li, A.~Rahmati, C.~Xiao, A.~Prakash,
  T.~Kohno, and D.~Song.
\newblock Robust physical-world attacks on deep learning visual classification.
\newblock In {\em The IEEE Conference on Computer Vision and Pattern
  Recognition (CVPR)}, June 2018.

\bibitem{finlayson2018adversarial}
S.~G. Finlayson, H.~W. Chung, I.~S. Kohane, and A.~L. Beam.
\newblock Adversarial attacks against medical deep learning systems.
\newblock {\em arXiv preprint arXiv:1804.05296}, 2018.

\bibitem{43405}
I.~Goodfellow, J.~Shlens, and C.~Szegedy.
\newblock Explaining and harnessing adversarial examples.
\newblock In {\em International Conference on Learning Representations}, 2015.

\bibitem{hariharan2011semantic}
B.~Hariharan, P.~Arbel{\'a}ez, L.~Bourdev, S.~Maji, and J.~Malik.
\newblock Semantic contours from inverse detectors.
\newblock In {\em 2011 International Conference on Computer Vision}, pages
  991--998. IEEE, 2011.

\bibitem{kurakin2016adversarial}
A.~Kurakin, I.~Goodfellow, and S.~Bengio.
\newblock Adversarial machine learning at scale.
\newblock {\em arXiv preprint arXiv:1611.01236}, 2016.

\bibitem{kurakin2018adversarial}
A.~Kurakin, I.~Goodfellow, S.~Bengio, Y.~Dong, F.~Liao, M.~Liang, T.~Pang,
  J.~Zhu, X.~Hu, C.~Xie, et~al.
\newblock Adversarial attacks and defences competition.
\newblock In {\em The NIPS'17 Competition: Building Intelligent Systems}, pages
  195--231. Springer, 2018.

\bibitem{meng2017magnet}
D.~Meng and H.~Chen.
\newblock Magnet: a two-pronged defense against adversarial examples.
\newblock In {\em Proceedings of the 2017 ACM SIGSAC Conference on Computer and
  Communications Security}, pages 135--147. ACM, 2017.

\bibitem{8237562}
J.~H. {Metzen}, M.~C. {Kumar}, T.~{Brox}, and V.~{Fischer}.
\newblock Universal adversarial perturbations against semantic image
  segmentation.
\newblock In {\em 2017 IEEE International Conference on Computer Vision
  (ICCV)}, pages 2774--2783, Oct 2017.

\bibitem{moosavi2016deepfool}
S.-M. Moosavi-Dezfooli, A.~Fawzi, and P.~Frossard.
\newblock Deepfool: a simple and accurate method to fool deep neural networks.
\newblock In {\em Proceedings of the IEEE conference on computer vision and
  pattern recognition}, pages 2574--2582, 2016.

\bibitem{papernot2016transferability}
N.~Papernot, P.~McDaniel, and I.~Goodfellow.
\newblock Transferability in machine learning: from phenomena to black-box
  attacks using adversarial samples.
\newblock {\em arXiv preprint arXiv:1605.07277}, 2016.

\bibitem{papernot2017practical}
N.~Papernot, P.~McDaniel, I.~Goodfellow, S.~Jha, Z.~B. Celik, and A.~Swami.
\newblock Practical black-box attacks against machine learning.
\newblock In {\em Proceedings of the 2017 ACM on Asia conference on computer
  and communications security}, pages 506--519, 2017.

\bibitem{papernot2016limitations}
N.~Papernot, P.~McDaniel, S.~Jha, M.~Fredrikson, Z.~B. Celik, and A.~Swami.
\newblock The limitations of deep learning in adversarial settings.
\newblock In {\em 2016 IEEE European Symposium on Security and Privacy
  (EuroS\&P)}, pages 372--387. IEEE, 2016.

\bibitem{papernot2016distillation}
N.~Papernot, P.~McDaniel, X.~Wu, S.~Jha, and A.~Swami.
\newblock Distillation as a defense to adversarial perturbations against deep
  neural networks.
\newblock In {\em 2016 IEEE Symposium on Security and Privacy (SP)}, pages
  582--597. IEEE, 2016.

\bibitem{samangouei2018defense}
P.~Samangouei, M.~Kabkab, and R.~Chellappa.
\newblock Defense-gan: Protecting classifiers against adversarial attacks using
  generative models.
\newblock {\em arXiv preprint arXiv:1805.06605}, 2018.

\bibitem{sitawarin2018darts}
C.~Sitawarin, A.~N. Bhagoji, A.~Mosenia, M.~Chiang, and P.~Mittal.
\newblock Darts: Deceiving autonomous cars with toxic signs.
\newblock {\em arXiv preprint arXiv:1802.06430}, 2018.

\bibitem{su2019one}
J.~Su, D.~V. Vargas, and K.~Sakurai.
\newblock One pixel attack for fooling deep neural networks.
\newblock {\em IEEE Transactions on Evolutionary Computation}, 23(5):828--841,
  2019.

\bibitem{42503}
C.~Szegedy, W.~Zaremba, I.~Sutskever, J.~Bruna, D.~Erhan, I.~Goodfellow, and
  R.~Fergus.
\newblock Intriguing properties of neural networks.
\newblock In {\em International Conference on Learning Representations}, 2014.

\bibitem{thys2019fooling}
S.~Thys, W.~Van~Ranst, and T.~Goedem{\'e}.
\newblock Fooling automated surveillance cameras: adversarial patches to attack
  person detection.
\newblock In {\em Proceedings of the IEEE Conference on Computer Vision and
  Pattern Recognition Workshops}, pages 0--0, 2019.

\bibitem{tramer2017ensemble}
F.~Tram{\`e}r, A.~Kurakin, N.~Papernot, I.~Goodfellow, D.~Boneh, and
  P.~McDaniel.
\newblock Ensemble adversarial training: Attacks and defenses.
\newblock {\em arXiv preprint arXiv:1705.07204}, 2017.

\bibitem{vincent2010stacked}
P.~Vincent, H.~Larochelle, I.~Lajoie, Y.~Bengio, and P.-A. Manzagol.
\newblock Stacked denoising autoencoders: Learning useful representations in a
  deep network with a local denoising criterion.
\newblock {\em Journal of machine learning research}, 11(Dec):3371--3408, 2010.

\bibitem{xie2017adversarial}
C.~Xie, J.~Wang, Z.~Zhang, Y.~Zhou, L.~Xie, and A.~Yuille.
\newblock Adversarial examples for semantic segmentation and object detection.
\newblock In {\em Proceedings of the IEEE International Conference on Computer
  Vision}, pages 1369--1378, 2017.

\bibitem{zhao2017pyramid}
H.~Zhao, J.~Shi, X.~Qi, X.~Wang, and J.~Jia.
\newblock Pyramid scene parsing network.
\newblock In {\em Proceedings of the IEEE conference on computer vision and
  pattern recognition}, pages 2881--2890, 2017.

\end{thebibliography}
}

\end{document}